# New SAR Target Recognition Based on YOLO and Very Deep Multi-Canonical Correlation Analysis

**Moussa Amrani**[*], **Abdelatif Bey**, and **Abdenour Amamra**
Ecole Militaire Polytechnique, Chahid Abderrahmane Taleb, Algiers, Algeria

**Abstract**. Synthetic Aperture Radar (SAR) images are prone to be contaminated by noise, which makes it very difficult to perform target classification in SAR images. Inspired by great success of very deep convolutional neural networks (CNNs), this paper proposes a robust feature extraction method for SAR image target classification by adaptively fusing effective features from different CNN layers. First, YOLOv4 network is fine-tuned to detect the targets from the respective MF SAR target images. Second, a very deep CNN is trained from scratch on the moving and stationary target acquisition and recognition (MSTAR) database by using small filters throughout the whole net to reduce the speckle noise. Besides, using small-size convolution filters decreases the number of parameters in each layer and therefore reduces computation cost as the CNN goes deeper. The resulting CNN model is capable of extracting very deep features from the target images without performing any noise filtering or pre-processing techniques. Third, our approach proposes to use the multi-canonical correlation analysis (MCCA) to adaptively learn CNN features from different layers such that the resulting representations are highly linearly correlated and therefore can achieve better classification accuracy even if a simple linear support vector machine (SVM) is used. Experimental results on the MSTAR dataset demonstrate that the proposed method outperforms the state-of-the-art methods.

## 1 Introduction

SYNTHETIC aperture radar (SAR) is a very high-resolution airborne and space-borne remote sensing system for imaging distant targets on a terrain, which can operate proficiently in all-weather day-and-night conditions and generate images of extremely high resolution. A SAR system sends electromagnetic pulses from radar mounted on a moving platform to a fixed particular area of interest on the target and combines the returned signals coherently to achieve a very high-resolution depiction of the scene.

SAR has been substantially employed for many applications such as surveillance, reconnaissance, and classification. However, speckle greatly disrupts SAR image readability, which makes defining discriminative and descriptive features a hard task. To address this problem, several feature processing methods have been advanced and utilized to understand the target from SAR images such as geometric descriptors, and transform-domain coefficients [1]. Although the above-mentioned methods may have some advantages, most of these methods are hand-designed and relatively simple [2]. Besides, they failed to achieve the promising classification performance (i.e., accuracy) [3]. Recently, deep convolutional neural networks (DCNNs) have been referenced by several authors to design classification algorithms for SAR images [4,5]. A deep neural network is an artificial network with multiple hidden layers between the input and the output. In [4] the authors proposed A-ConvNets which consists of sparsely connected layers to reduce the number of free parameters from over-fitting. After implementing data augmentation, the network is trained using mini-batch stochastic gradient descent with momentum and back-propagation algorithm. Then, a soft-max layer is applied to output a probability distribution over class labels. One of the uncertain points of this method is that the convolutional layers suffer less from over-fitting because they have smaller number of parameters compared to the number of activations. Therefore, adding dropout to convolutional layers slows down the training [5]. In addition, they employed cropped patches of $88 \times 88$ from the original SAR images in the training phase (i.e., as data augmentation) to deal with the translation invariance of DCNN for SAR-ATR system [6]. The other well-known deep learning based algorithm [7] focuses on solving the classification problem by learning randomly sampled image patches using unsupervised sparse auto-encoder instead of using the classical back-propagation algorithm. Then a single layer of convolutional neural network is used to automatically learn features from SAR images. These feature maps are then adopted to train the final soft-max classifier, which results in a little low classification accuracy [8]. The research in [9], proposes to select deep features for SAR target classification task, in which the top layers of CNNs contain more semantic information and describe the global features of the images, whereas the intermediate layers describe the local features, and the bottom layers contain more low-level information for the description of texture, edges, etc. However, the authors used large kernels in the first and the second layers, which increase the number of parameters and have less discriminative decision functions. Moreover, they adopted a pre-processing method to remove some noise from target images, and metric learning for feature selection to adjust the accuracy level, which is time consuming.

More recently, very deep CNNs using small-size convolution filters have been used by Ciresan et al [10]. However, their networks are significantly less deep and they have not been evaluated on SAR images. Goodfellow et al. [11] applied very deep ConvNets (11 weight layers) to street number recognition tasks and have depicted that increasing the depth of the network produces better performance. Szegedy et al. [12], an ILSVRC-2014 top-performing classification task, has developed independently a network based on very deep ConvNets (22 weight layers) and small convolution filters. However, their network is more complex and the spatial resolution of the feature maps in the first layers is sharply reduced to decrease the computation amount [13]. C.P. Schwegmann et al. [14] have introduced very deep features for ship discrimination in SAR images.

When trained on a small database, their proposed very deep high network can provide better classification performance than conventional DCNNs.

Compared to the handcrafted and deep features, our approach uses very deep features to have more powerful discriminative and robust representation abilities, and multi-canonical correlation analysis (MCCA) algorithm to maximize the correlation among the feature sets, remove irrelevant features and overcome the curse of dimensionality issue. The contributions of the proposal method mainly include two aspects.

• YOLOv4 is used for SAR target detection. Besides, a very deep network is proposed for SAR image target classification, which uses very small receptive small filter sizes throughout the whole net to reduce the speckle noise that degrades the quality of SAR images, and therefore achieves better performance as we go deeper.

• Multi-Canonical Correlation Analysis (MCCA) is proposed to adaptively select and fuse CNN features from different layers and such that the resulting representations are highly linearly correlated and therefore remove irrelevant features, speed up the training task, and improve the classification accuracy. As a result, we come up with **particularly** more accurate CNN-SAR architecture, which achieves the state-of-the-art accuracy on **MSATR SAR target** recognition tasks.

 even when adopted as a relatively simple pipeline.

The rest of this paper is organized as follows. Section 2 describes the framework of the proposed method and introduces the feature fusion method using MCCA. The experimental results are presented in Section 3. Finally, Section 4 gives the concluding remarks of the paper.

## 2    YOLOv4 Target detection

YOLOv4 combines many advanced target detection techniques to improve the accuracy and running speed of CNN as it is clarified in Figure 1. It can be run in real-time on a typical GPU, which makes it widely used. Yolo V4 consists of four parts. The main methods and tricks utilized in each part are as follows [15]: **Input:** Mosaic data augmentation, cross mini batch normalization (CmBN), and self-adversarial training (SAT). **BackBone:** Cross stage partial connections Darknet53 (CSPDarknet53), mish-activation, and dropblock regularization. **Neck:** SPP, modified feature pyramid network (FPN), path aggregation network (PAN). Prediction: Modified complete IOU (C-IOU) loss, distance IOU (D-IOU) nms. Some of these tricks can obviously improve the network performance. Mosaic data augmentation mixes four training images with clipping and scaling, which significantly reduces the requirement for large mini-batch processing and GPU computing. SAT alters the original SAR image to create the deception of undesired object. These modified images are used for network training in next stage, which could improve the robustness of the network. Cross stage partial connections split the gradient flow propagate into different network paths, which greatly reduces the amount of computation, and improves the inference speed of the network. Other tricks also improve the other details of the networks. For example, SPP, FPN and PAN improve the feature extraction. Modified C-IOU loss and DIOU nms improve the IOU loss to achieve better convergence speed and accuracy of regression problem.

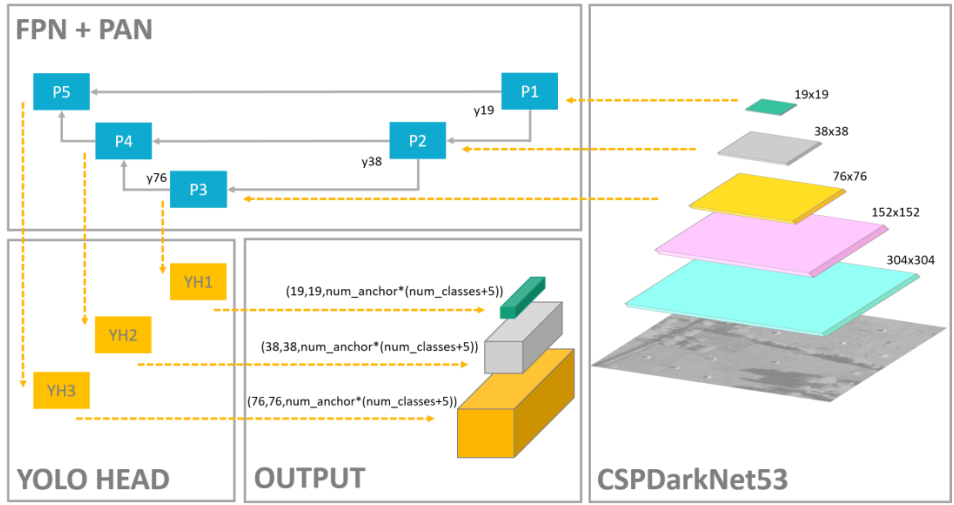

**Fig.1**. The architecture of YOLOv4

## 3    The Proposed Method for SAR Target Recognition

Inspired by great success of very deep convolutional neural networks (VDCNNs), this paper presents a robust feature extraction method for SAR target classification using very deep features without performing any noise filtering or pre-processing techniques. Figure 2 shows the main steps of the proposed network: after targets detection from the respective SAR target images, SAR Oriented Network (SARON) is used for extracting very deep features from detected data. Then the resulting deep features are then selected and fused by MCCA. Finally, the classification of the targets with respect to its training set is done according to the classification error rates using SVM. In the following sections, each step of the proposed method is described in detail.

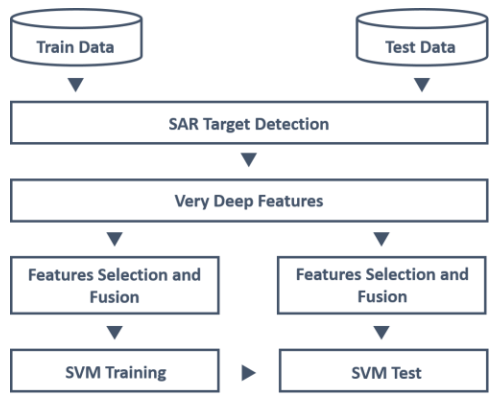

**Fig.2**. Overall architecture of the proposed method

*3.1 SAR Oriented Network*

Using deep neural networks to learn effective feature representations has become popular in SAR target classification [4, 5, 16, 17]. It is also employed in the military field as automatic target recognition (ATR) [18]. In contrast to most DCNNs that usually have five or seven layers, the proposed network is based on the very deep VGG net for feature extraction, which has a much deeper architecture (up to 19 weight layers) and hence can provide much informative and descriptive features [13]. To achieve better classification performance, the network is fine-tuned from scratch on the MSTAR database and then the fine-tuned network is treated as a fixed feature descriptor for SAR target images.

Figure 3 shows the architecture of the proposed very deep CNN model, which is a stack of convolutional layers (Conv.) followed by three fully-connected (Fc.) layers, and each stack of Conv. is followed by a max pooling layer (Pool.). Besides, all hidden layers are supplied with ReLU. Fc.1 is regularized using dropout technique, while the last layer acts as a C-class SVM classifier. Figure 4 shows the outputs from the first convolutional layer corresponding to a sample SAR image from the MSTAR dataset.

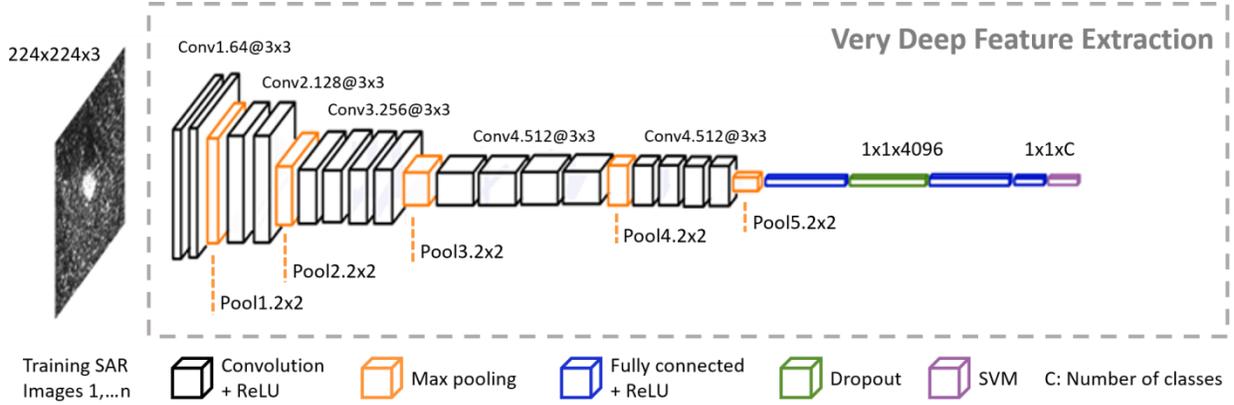

**Fig.3**.The proposed SAR oriented very deep network for target image classification

*3.1.1 Configuration*

The network used in this paper has 19 weight layers (16 conv. and 3 Fc. layers). The width of the convolutional layers (channels number) is quite small, which begins from 64 in the first layer, then increases by a factor of 2 after each max pooling layer until it reaches 512.

Our Network configuration is rather different from the ones used in the previous MSTAR SAR target classification [4, 9, 15, 17]. Rather than using relatively large receptive fields in the first conv. layers (e.g.13×13 with stride 4 in [15]), 7×7 with stride 1 in [9], or 5×5 with stride 1 in [4]), we use very small $3 \times 3$ receptive fields throughout the whole net, which are convolved with the input at every pixel (with stride 1). Furthermore, we remove local response normalization (LRN) as it does not improve the performance on our SAR images dataset but leading to increase the computation time and the memory consumption, and we modify the output layers to appropriately match the *C* classes in our data set. The configuration of the fully connected layers is the same in all network and all hidden layers are supplied with the rectified linear unit (ReLU) to have more discriminative decision functions and lower computation cost [13, 19]. To achieve better classification accuracy, the linear support vector machine (*L*2-SVM) is used as a baseline classifier instead of maximum entropy classifier [18, 20]. Consequently, we come up with expressively more accurate CNN architecture, which achieves the state-of-the-art accuracy on SAR target classification task.

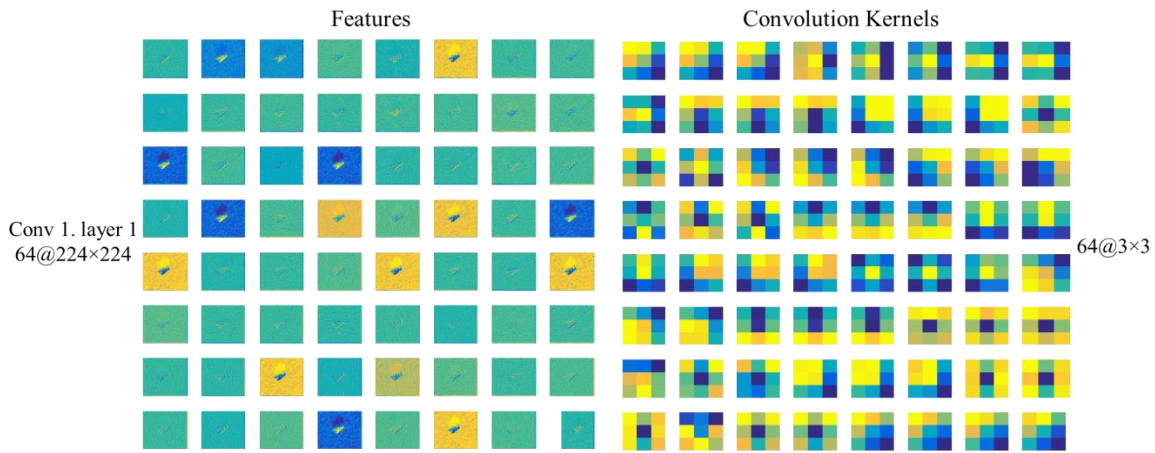

**Fig.4**. (Left) Output feature maps and (right) Learned kernels of the first layer.

*3.2 MCCA Based Feature level Fusion*

Our proposed method is based on the fusion of the very deep features learned by our network model. The outputs of some selected layers are used as a feature descriptor of the input target. In particular, we considered the fully connected layers output with 4096 feature channels, which contain more low-level information for the description of texture and edges.

Several feature fusion methods have been proposed to obtain more informative descriptors to describe the image targets. The serial strategy [21] simply combines two feature sets into one real union-vector and assumes that $x$ and $y$ are two feature sets with $p$ and $q$ vector dimensions, respectively. Contrary to the serial feature fusion, the parallel strategy [22] combines the two feature sets into a complex feature set $z = x + iy$ where $i$ is the imaginary unit. However, these strategies neglect the class structure among the sample images and attain high dimension of the fused feature sets. Besides, SAR images have their own characteristics (speckle noise, illumination, pose variations, depression angle, corruption, and occlusion), and the choice of features is dictated by the statistical structure of the data [23].

Let $X_{(p \times n)}$ and $Y_{(q \times n)}$ be two matrices that contain $n$ training feature vectors. In our study, we adopt CCA [24, 25] to find the linear combination $a^T$ and $b^T$ that maximize the pairwise correlations $X^* = a^T X$ and $Y^* = b^T Y$ across the two feature sets, restrict the correlations to be between classes, and to face noise [26]. The top layers of our proposed SAR-OVDN contain more semantic information and describe the global feature of the images, whereas the intermediate layers describe the local features, and the bottom layers contain more low-level information for the description of texture, edges. MCCA is proposed to combine information from different levels to obtain distinguishing (relevant) features and overcome the curse of dimensionality problem. The transformation matrices $a$ and $b$ are diagonalized by solving the eigenvalue equations as [27]:

$$V = \begin{pmatrix} \text{cov}(x) & \text{cov}(x,y) \\ \text{cov}(y,x) & \text{cov}(y) \end{pmatrix} = \begin{pmatrix} V_{xx} & V_{xy} \\ V_{Yx} & V_{yy} \end{pmatrix} \quad (4)$$

$$\begin{cases} V_{xx}^{-1} V_{xy} V_{yy}^{-1} V_{yx} \hat{a} = \Lambda^2 \hat{a} \\ V_{yy}^{-1} V_{yx} V_{xx}^{-1} V_{xy} \hat{b} = \Lambda^2 \hat{b} \end{cases} \quad (5)$$

where $\hat{a}$ and $\hat{b}$ are the eigenvectors and $\Lambda^2$ is the eigenvalues diagonal matrix, the non-zero eigenvalues number in each equation is $r = rank\ (V_{xy}) \leq \min\ (n, p, q)$ that is organized in decreasing order, $\gamma_1 \geq \gamma_2 \geq \cdots \geq \gamma_r$, the transformation matrices $a$ and $b$ contain the consistent eigenvectors to the non-zero eigenvalues, and $X^*, Y^* \in \Re^{r \times n}$ are the canonical variates. To find the transformed training features sets, the covariance matrix in Equation 4 will be defined as:

$$V^* = \begin{pmatrix} 1 & 0 & \cdots & 0 & \gamma_1 & 0 & \cdots & 0 \\ 0 & 1 & \cdots & 0 & 0 & \gamma_2 & \cdots & 0 \\ \vdots & & \ddots & & \vdots & & \ddots & \\ 0 & 0 & \cdots & 1 & 0 & 0 & \cdots & \gamma_r \\ \gamma_1 & 0 & \cdots & 0 & 1 & 0 & \cdots & 0 \\ 0 & \gamma_2 & \cdots & 0 & 0 & 1 & \cdots & 0 \\ \vdots & & \ddots & & \vdots & & \ddots & \\ 0 & 0 & \cdots & \gamma_r & 0 & 0 & \cdots & 1 \end{pmatrix} \quad (6)$$

where the canonical variates have nonzero correlation only on their relative indices. The upper left and lower right corners in the identity matrices indicate that the canonical variates are uncorrelated with in training features sets. Figure 5 shows the statistical transformation of the CCA framework.

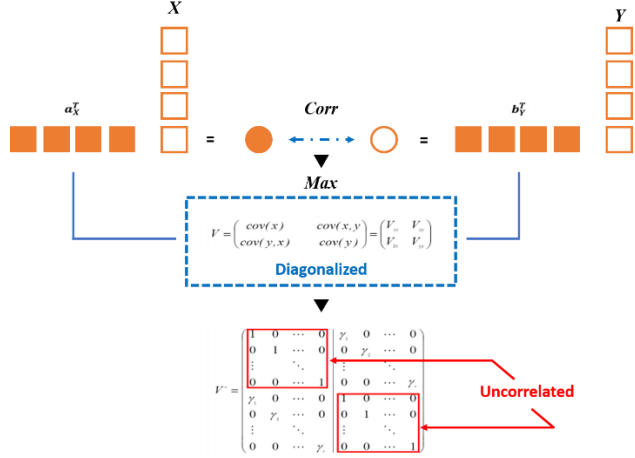

**Fig.5**. CCA transformation framework

In our work, feature-level fusion is then performed by using summation of the transformed feature vectors as:

$$M = X^* + Y^* = a^T X + b^T Y = \begin{pmatrix} a \\ b \end{pmatrix}^T \begin{pmatrix} X \\ Y \end{pmatrix} \quad (7)$$

where *M* is the Very Deep Canonical Correlation Discriminant Features (VDCCDFs). Multi-canonical correlation analysis (MCCA) generalizes CCA to be appropriate for more than two features sets. We suppose that we have $\lambda$ feature sets $F_i \in \Re^{p_i \times n}$, $i = 1, 2, \ldots, \lambda$, which are arranged based on their rank, rank($F_1$) $\geq$ rank($F_2$) $\geq \ldots \geq$ rank($F_k$). MCCA applies CCA to two features sets at the same time. In each phase, the two feature sets with the highest ranks are fused together to keep the maximum possible feature vectors length as shown in Figure 6.

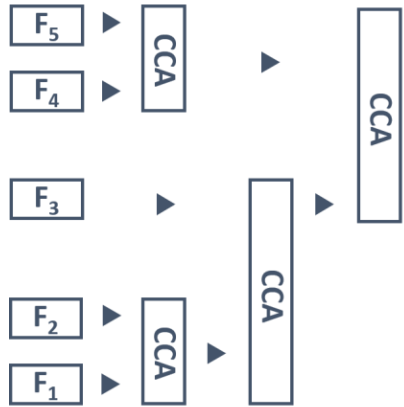

**Fig.6**. Multi-canonical correlation analysis techniques for five sample sets with rank ($F_1$) > rank ($F_2$) > rank ($F_3$) > rank ($F_4$) = rank ($F_5$)

*3.3 Support Vector Machine (L2-SVM)*

Recently, most of the deep learning models utilize multi-class logistic regression for prediction and minimize cross-entropy loss for SAR target classification [4, 5, 16, 28]. In this paper, for better performance and parameter optimization, *L*2-SVM [20] is adopted to train our SAR Oriented Very deep CNN model, when the learning minimizes a margin-based loss by back propagating the gradients from the top layer linear SVM. Therefore, we differentiate the SVM objective function with respect to the activation of the penultimate layer as follows. Let $x_n \in \Re^D$ be training feature set and $y_n$ its labels, where $n=1,..., N$, and $t_n \in \{-1,+1\}$. Given the objective $L(w)$ in Equation 8, and the input $x$ is replaced with the penultimate activation $m$ as:

$$\min_{w} \frac{1}{2} w^T w + C \sum_{n=1}^{N} \max(1 - w^T x_n t_n, 0) \quad (8)$$

$$\frac{\partial L(w)}{\partial m_n} = -C t_n w \left( II\{1 > w^T m_n t_n\} \right) \quad (9)$$

where $II\{.\}$ is the indicator function and $C$ is a constant ($C \geq 0$). As before, for the *L*2-SVM, we have:

$$\frac{\partial l(w)}{\partial m_n} = -2 C t_n w (\max(1 - w^T m_n t_n, 0)) \quad (10)$$

Based on this point, the back-propagation algorithm is just the same as the standard soft-max for deep learning networks.

## 4  Experimental Results and Analysis

The results of SAR target detection based on the MF dataset, and to validate the effectiveness of the proposed method we conduct extensive experiments on two MSTAR datasets, which are publically available [29]. In the following sections, we first describe the implementation details, and then introduce the used datasets. Finally, the experimental results are presented and analyzed.

*4.1 Implementation details*

The pre-trained AlexNet, CaffeRef, VGGs and Very Deep-16 are fine-tuned on real SAR images from MSTAR database. The proposed network SARON is trained on a 2.7-GHz CPU with 64 GB of memory and a moderate graphics processing unit (GPU) card. All methods have been implemented using Microsoft Windows 10 Pro 64-bit and MATLAB R2016a. We randomly select the SAR image samples for the training and the testing datasets.

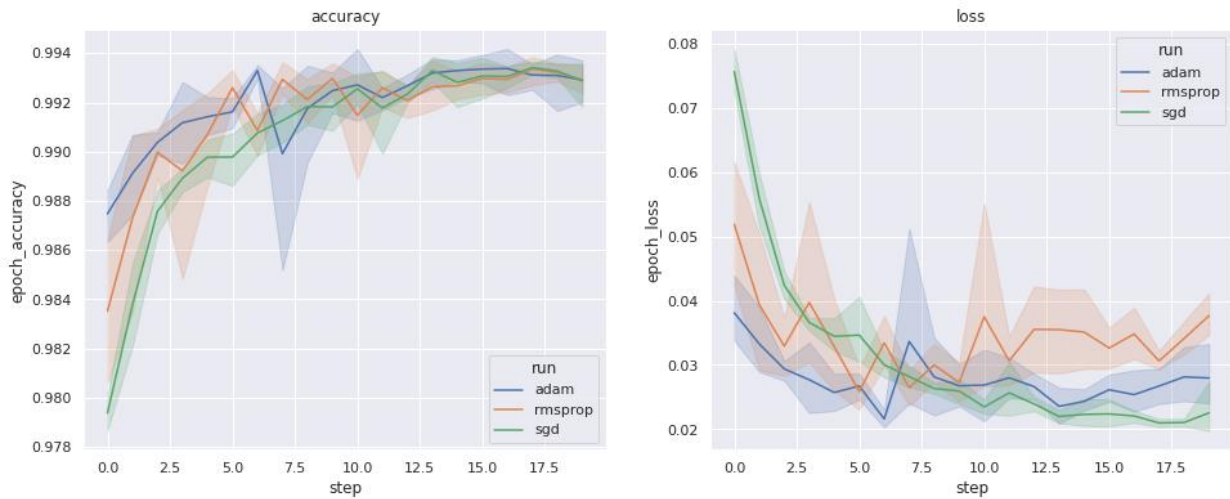

**Fig.7**. The effect of optimization methods on the performance.

To study the effect of adaptive optimization algorithms on the accuracy in SAR image recognition system, several successful methods such as Stochastic Gradient Descent (sgd), Adam, RMSPROP are evaluated in our work as it is depicted in Figure 7, where the training is carried out by optimizing the primal L2-SVM objective to learn lower level features, the batch size is set to 64, a momentum parameter to 0.9. The training is regularized by weight decay ($L_2$ penalty multiplier set to $5 \times 10^{-4}$) and dropout regularization for the first fully-connected layer (dropout ratio set to 0.5). The learning rate is initially set to $10^{-2}$, and then is decreased by a factor of 10 when the accuracy of the validation set stopped improving. In total, the learning rate was decreased 2 times, and the learning has been stopped after 73 epochs. The classification of the targets with respect to its training set is done according to the classification error rates using the linear SVM. The main advantages of our proposed method as follows: first, using small-size convolution throughout the whole net to reduce the speckle noise. Second, decreases the number of parameters in each layer and therefore reduces computation cost as the CNN goes deeper. Moreover, ReLU layers are integrated to have more discriminative decision functions and lower computation cost. Third, the MCCA algorithm is utilized to fuse the feature vectors from the fully connected layers by summation and concatenation forming new robust and discriminant features. The analysis of the advantages is discussed in the following subsections.

*4.2 Datasets*

The SAR images used in experiments are from the Moving and Stationary Target Acquisition and Recognition (MSTAR) database. This benchmark data is acquired by the Sandia National Laboratories Twin Otter SAR sensor payload, operating at X-band with a high resolution of 0.3 m, spotlight mode, and HH single 320 polarizations (i.e., single-channel): where the phase content is entirely discarded because it is random and uniformly distributed [30, 31]. The first dataset is the MSTAR public mixed target dataset which includes ten military vehicle targets: (armored personnel carrier: BMP-2, BRDM-2, BTR-60, and BTR-70; tank: T-62, T-72; rocket launcher: 2S1; air defense unit: ZSU-234; truck: ZIL-131; bulldozer: D7). The optical images and their relative SAR images are shown in Figure 8.

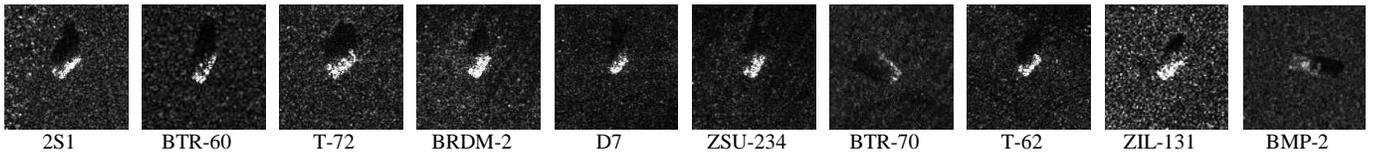

2S1    BTR-60    T-72    BRDM-2    D7    ZSU-234    BTR-70    T-62    ZIL-131    BMP-2

**Fig.8**. Types of military targets: optical images in the top associated with their relative SAR images in the bottom

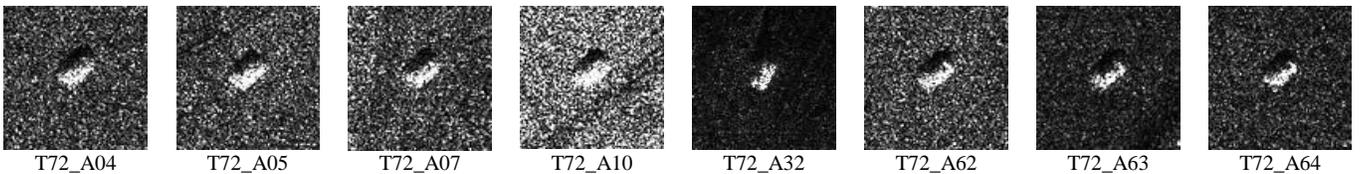

T72_A04    T72_A05    T72_A07    T72_A10    T72_A32    T72_A62    T72_A63    T72_A64

**Fig.9.** T-72 multi variants: optical images in the top associated with their relative SAR images in the bottom.

The second dataset chosen for evaluation is the MSTAR public T-72 Variants dataset, which contains eight T-72variants: A04, A05, A07, A10, A32, A62, A63, and A64. Optical images and the corresponding SAR images of the eight T-72 targets are shown in Figure 9.

*4.3 Results on the MSTAR Public Mixed Target Dataset*

The dataset details including depression angles with target signatures of all MSTAR images used in this task are listed in Table 1. BMP-2 and BTR-70 refers to man-made (metal) objects armored personnel carrier targets, and T-72 refers to main battle tank. The single and overall accuracies are well explained in Tables 2 and 3, respectively. The confusion matrix is clarified in Figure 10.

TABLE 1 MSTAR PUBLIC MIXED TARGET DATASET: THE NUMBER OF TRAINING AND TESTING SAMPLES USED IN THE EXPERIMENTS

| Target | | Train | | Test | |
|---|---|---|---|---|---|
| | | Depression | Number of Images | Depression | Number of Images |
| BMP-2 | SN-C21 | 17° | 233 | 15° | 196 |
| | SN-9563 | 17° | 233 | 15° | 195 |
| | SN-9566 | 17° | 232 | 15° | 196 |
| BTR-70 | | 17° | 233 | 15° | 196 |
| T-72 | SN-132 | 17° | 232 | 15° | 196 |
| | SN-812 | 17° | 231 | 15° | 195 |
| | SN-S7 | 17° | 228 | 15° | 191 |
| BTR-60 | | 17° | 256 | 15° | 196 |
| 2S1 | | 17° | 299 | 15° | 274 |
| BRDM-2 | | 17° | 299 | 15° | 274 |
| D7 | | 17° | 299 | 15° | 274 |
| T-62 | | 17° | 299 | 15° | 274 |
| ZIL-131 | | 17° | 299 | 15° | 274 |
| ZSU-234 | | 17° | 299 | 15° | 274 |

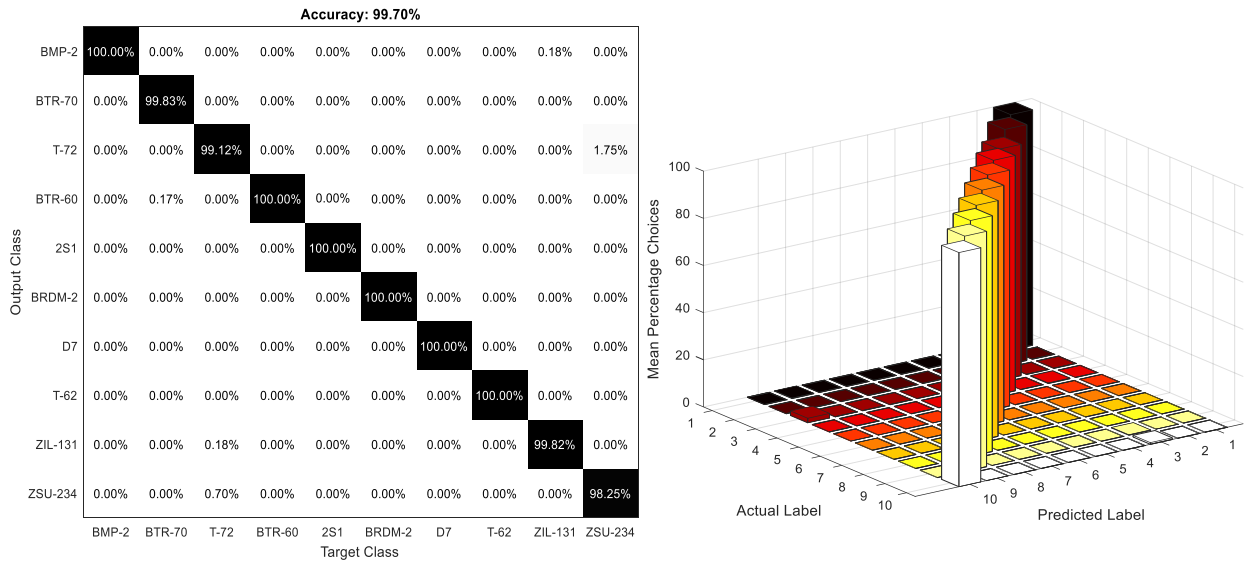

**Fig.10**. Confusion matrix of the proposed method on MSTAR public mixed target dataset. The rows and columns of the matrix indicate the actual and predicted classes, respectively.

The noise in SAR images can be multiplicative or additive [32, 33]. Figure 11 shows the noise simulation scheme, where the value of a randomly selected pixel is replaced by a value from a uniform distribution. The anti-noise performance of the proposed method is compared with the noise simulation paradigm in [4, 34] and shows that our proposed method is more robust to noise corruption as clarified in Table 4.

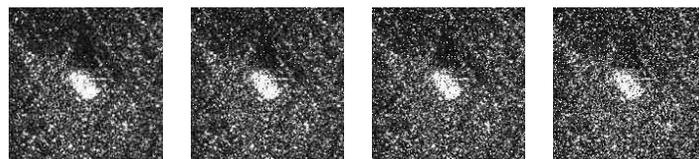

**Fig. 11.** Illustration of random noise. The levels of noise are 1%, 5%, 10%, and 15%, respectively

TABLE 4 ACCURACIES ACROSS THE LEVEL OF NOISE

| Noise | 1% | 5% | 10% | 15% |
|---|---|---|---|---|
| A-Convnets [4] | 0.9176 | 0.8852 | 0.7584 | 0.5468 |
| SRC [34] | 0.9276 | 0.8639 | 0.8049 | 0.6419 |
| SVM [34] | 0.8826 | 0.8451 | 0.5310 | 0.4242 |
| KNN [34] | 0.9341 | 0.8579 | 0.8177 | 0.5573 |
| MSRC [34] | 0.9488 | 0.8739 | 0.8476 | 0.6853 |
| SAR-Oriented GBVS [35] | 0.9216 | 0.8922 | 0.7609 | 0.5578 |
| The proposed method | **0.9517** | **0.9012** | **0.8529** | **0.6998** |

*4.4 Results on the MSTAR Public T-72 Variants Dataset*

In another scenario, the proposed method is evaluated on the more challenging T-72 eight-target classification problem, as all the targets are almost indistinguishable. The number and the depression angles of the training and the testing sample images are listed in Table 5. The single and overall accuracies are explained in Tables 6 and 7, respectively. The confusion matrix is presented in Figure 12.

TABLE 5 THE MSTAR PUBLIC T-72 VARIANTS DATASET: THE NUMBER AND THE DEPRESSION ANGLES OF TRAINING AND TESTING SAMPLES USED IN THE EXPERIMENTS.

| Target | Train | | Test | |
|---|---|---|---|---|
| | Depression | Number of Images | Depression | Number of Images |
| A04 | 17° | 299 | 15° | 274 |
| A05 | 17° | 298 | 15° | 274 |
| A07 | 17° | 299 | 15° | 274 |
| A10 | 17° | 296 | 15° | 271 |
| A32 | 17° | 298 | 15° | 274 |
| A62 | 17° | 299 | 15° | 274 |
| A63 | 17° | 299 | 15° | 273 |
| A64 | 17° | 298 | 15° | 274 |

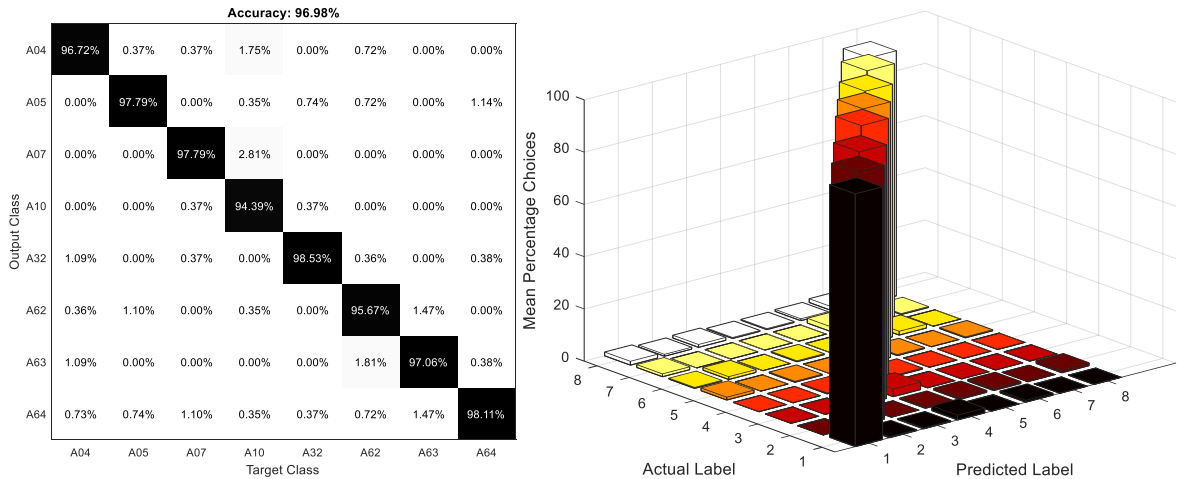

**Fig.12.** Confusion matrix of the proposed method on T-72 Variants dataset. The rows and columns of the matrix indicate the actual and predicted classes, respectively.

## 4.5 Results on the MSTAR with Large Depression Angle Variations

Since SAR images are very sensitive to depression angle variation, the credibility of the proposed method is further evaluated on large depression angles. In these experiments, four targets (2S1, BRDM-2, T-72, and ZSU-234) with 30° depression angle are assessed. The types and the number of the considered SAR images are shown in Table 8.

TABLE 8 THE MSTAR PUBLIC DATABASE WITH 30° DEPRESSION ANGLE: THE NUMBER OF TRAINING AND TESTING SAMPLES USED IN THE EXPERIMENTS.

| Target | Type | Depression | Number of Images |
|---|---|---|---|
| 2S1 | B01 | 30° | 288 |
| BRDM-2 | E71 | 30° | 287 |
| T-72 | A64 | 30° | 288 |
| ZSU-234 | D08 | 30° | 288 |

The single and overall accuracies are presented in Tables 9 and 10, respectively. The confusion matrix is shown in Figure 13.

## 4.6 Comparison with Different Fusion Methods

In order to study the sensitivity of the feature level fusion on classification on the MSTAR database, we compared the classification accuracy with different fusion strategies (CCA, MCCA), SAR-Oriented GBVS [35] and without feature fusion. As shown in Table 11, our proposed method using MCCA achieves the best performance.

TABLE 2 THE CLASSIFICATION ACCURACY FOR EACH TARGET CLASS ON MSTAR PUBLIC MIXED TARGET DATASET

| Class | Single Accuracy | Error Single | Total Accuracy | Error Total | Sensitivity | Specificity | Precision | False Positive Rate |
|---|---|---|---|---|---|---|---|---|
| BMP-2 | 0.99825 | 0.0017452 | 0.10223 | 0 | 0.99825 | 1 | 1 | 0 |
| BTR-70 | 1 | 0 | 0.10223 | 0.00017873 | 1 | 0.9998 | 0.99825 | 0.00019908 |
| T-72 | 0.98255 | 0.017452 | 0.10063 | 0.00089366 | 0.98255 | 0.999 | 0.9912 | 0.00099562 |
| BTR-60 | 0.99778 | 0.0022173 | 0.080429 | 0 | 0.99778 | 1 | 1 | 0 |
| 2S1 | 1 | 0 | 0.10241 | 0 | 1 | 1 | 1 | 0 |
| BRDM-2 | 1 | 0 | 0.10223 | 0 | 1 | 1 | 1 | 0 |
| D7 | 1 | 0 | 0.10223 | 0 | 1 | 1 | 1 | 0 |
| T-62 | 1 | 0 | 0.10241 | 0 | 1 | 1 | 1 | 0 |
| ZIL-131 | 0.99824 | 0.0017575 | 0.10152 | 0.00017873 | 0.99824 | 0.9998 | 0.99824 | 0.00019897 |
| ZSU-234 | 0.99295 | 0.0070547 | 0.10063 | 0.0017873 | 0.99295 | 0.99801 | 0.98255 | 0.0019889 |

TABLE 3 THE OVERALL CLASSIFICATION ACCURACY MSTAR PUBLIC MIXED TARGET DATASET

| Accuracy | Error | Sensitivity | Specificity | Precision | False Positive Rate |
|---|---|---|---|---|---|
| **0.9970** | 0.0030 | 0.9970 | 0.9997 | 0.9970 | 3.3825e-04 |

TABLE 6 THE CLASSIFICATION ACCURACY FOR EACH TARGET CLASS ON MSTAR PUBLIC T-72 VARIANTS DATASET

| Class | Single Accuracy | Error Single | Total Accuracy | Error Total | Sensitivity | Specificity | Precision | False Positive Rate |
|---|---|---|---|---|---|---|---|---|
| A04 | 0.96715 | 0.032847 | 0.12112 | 0.0041133 | 0.96715 | 0.9953 | 0.96715 | 0.0047022 |
| A05 | 0.9708 | 0.029197 | 0.12157 | 0.0027422 | 0.9708 | 0.99687 | 0.97794 | 0.0031348 |
| A07 | 0.9708 | 0.029197 | 0.12157 | 0.0027422 | 0.9708 | 0.99687 | 0.97794 | 0.0031348 |
| A10 | 0.99262 | 0.0073801 | 0.12294 | 0.0073126 | 0.99262 | 0.99165 | 0.94386 | 0.0083464 |
| A32 | 0.9781 | 0.021898 | 0.12249 | 0.0018282 | 0.9781 | 0.99791 | 0.98529 | 0.0020899 |
| A62 | 0.96715 | 0.032847 | 0.12112 | 0.0054845 | 0.96715 | 0.99373 | 0.95668 | 0.0062696 |
| A63 | 0.96703 | 0.032967 | 0.12066 | 0.0036563 | 0.96703 | 0.99582 | 0.97059 | 0.0041775 |
| A64 | 0.94526 | 0.054745 | 0.11837 | 0.0022852 | 0.94526 | 0.99739 | 0.98106 | 0.0026123 |

TABLE 7 THE OVERALL CLASSIFICATION ACCURACY ON MSTAR PUBLIC T-72 VARIANTS DATASET

| Accuracy | Error | Sensitivity | Specificity | Precision | False Positive Rate |
|---|---|---|---|---|---|
| **0.9698** | 0.0302 | 0.9699 | 0.9957 | 0.9701 | 0.0043 |

TABLE 9 THE CLASSIFICATION ACCURACY FOR EACH TARGET CLASS ON MSTAR PUBLIC DATABASE WITH 30° DEPRESSION ANGLE

| Class | Single Accuracy | Error Single | Accuracy in Total | Error in Total | Sensitivity | Specificity | Precision | False Positive Rate |
|---|---|---|---|---|---|---|---|---|
| 2S1 | 0.98958 | 0.010417 | 0.24761 | 0.0026064 | 0.98958 | 0.99652 | 0.98958 | 0.0034762 |
| BRDM-2 | 0.98955 | 0.010453 | 0.24674 | 0.004344 | 0.98955 | 0.99421 | 0.9827 | 0.005787 |
| T-72 | 0.98264 | 0.017361 | 0.24587 | 0.0026064 | 0.98264 | 0.99652 | 0.98951 | 0.0034762 |
| ZSU-234 | 0.99306 | 0.0069444 | 0.24848 | 0.0017376 | 0.99306 | 0.99768 | 0.99306 | 0.0023175 |

TABLE 10 THE OVERALL CLASSIFICATION ACCURACY ON MSTAR PUBLIC DATABASE WITH 30° DEPRESSION ANGLE

| Accuracy | Error | Sensitivity | Specificity | Precision | False Positive Rate |
|---|---|---|---|---|---|
| **0.9887** | 0.0113 | 0.9887 | 0.9962 | 0.9887 | 0.0038 |

TABLE 11 COMPARISON WITH DIFFERENT FUSION METHODS ON MSTAR DATABASE

| Method | MSTAR public mixed target (%) | MSTAR public T-72 Variants (%) | MSTAR with 30° (%) |
|---|---|---|---|
| Proposed without fusion | 98.79 | 94.50 | 96.13 |
| VDCCA | 99.12 | 95.31 | 97.55 |
| VDMCCA | 99.70 | 96.98 | 98.87 |
| SAR-Oriented GBVS [35] | 99.68 | 96.90 | 98.70 |

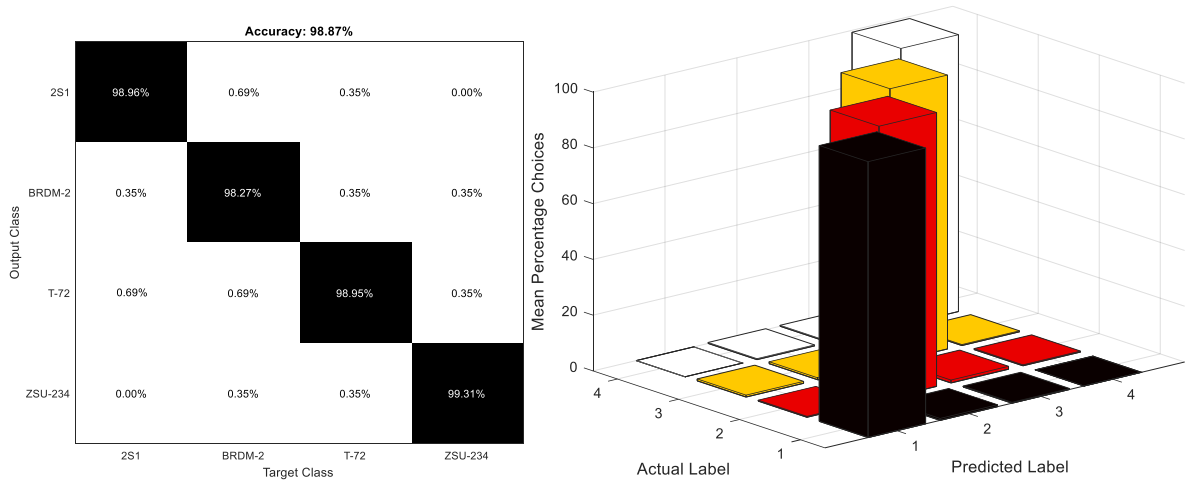

**Fig.13.** Confusion matrix of the proposed method on MSTAR with large depression angle variations. The rows and columns of the matrix indicate the actual and predicted classes, respectively.

*4.7 Influence of Number of Layers and Different Receptive Fields on the Results*

To study the effect of network depth on its accuracy in SAR image recognition setting, several successful CNN models pre-trained on MSTAR are evaluated in our work, which are the famous baseline model AlexNet, the Caffe reference model (CaffeRef), and the VGG network [36]. Due to a large learning capacity, dominant expressive power and hierarchical structure of our SAR oriented very deep network (19 layers): a high-level, semantic and robust feature representation for each region proposal is obtained as it is illustrated in Table 12.

TABLE 12 THE EFFECT OF NETWORK DEPTH ON THE PERFORMANCE

| | Accuracy (%) | | | | | | | | MSTAR public mixed | MSTAR T-72 Variants | MSTAR depression 30° |
|---|---|---|---|---|---|---|---|---|---|---|---|
| Models | Conv1 | Conv2 | Conv3 | Conv4 | Conv5 | Fc1 | Fc2 | Fc3 | | | |
| AlexNet | 11×11×96 | 5×5×256 | 3×3×384 | 3×3×384 | 3×3×256 | 4096 dropout | 4096 dropout | $C$ softmax | 95.21 | 86.82 | 89.66 |
| CaffeRef | 11×11×96 | 5×5×256 | 3×3×384 | 3×3×384 | 3×3×256 | 4096 dropout | 4096 dropout | $C$ softmax | 95.45 | 85.96 | 88.99 |
| VGGF | 11×11×64 | 5×5×256 | 3×3×256 | 3×3×256 | 3×3×256 | 4096 dropout | 4096 dropout | $C$ softmax | 95.66 | 86.9 | 90.6 |
| VGGM | 7×7×96 | 5×5×256 | 3×3×512 | 3×3×512 | 3×3×512 | 4096 dropout | 4096 dropout | $C$ softmax | 96.7 | 88.08 | 91.5 |
| VGGM-128 | 7×7×96 | 5×5×256 | 3×3×512 | 3×3×512 | 3×3×512 | 4096 dropout | 128 dropout | $C$ softmax | 96.13 | 89.43 | 90.12 |
| VGGM-1024 | 7×7×96 | 5×5×256 | 3×3×512 | 3×3×512 | 3×3×512 | 4096 dropout | 1024 dropout | $C$ softmax | 97.6 | 90.86 | 91.6 |
| VGGM-2048 | 7×7×96 | 5×5×256 | 3×3×512 | 3×3×512 | 3×3×512 | 4096 dropout | 2048 dropout | $C$ softmax | 96.74 | 90.12 | 91.55 |
| VGGS | 7×7×96 | 5×5×256 | 3×3×512 | 3×3×512 | 3×3×512 | 4096 dropout | 4096 dropout | $C$ softmax | 97.95 | 91.66 | 92.88 |
| *VGG16* | *16 weight layers including 13 convolutional layers 3×3 and 3 fully-connected layers* | | | | | | | | 99.1 | 95.39 | 97.1 |
| *SAR-OVDN (19 weight layers including 16 convolutional layers and 3 fully-connected layers)* | | | | | | | | | **99.7** | **96.98** | **98.87** |

In addition, using a stack of two 3 × 3 convolutional layers (without pooling in between) has a 5 × 5 effective receptive field, and using three of 3 × 3 layers has a 7 × 7 effective receptive field. So the reasons for using, for instance, a stack of three 3×3 convolutional layers instead of a single 7×7 layer are: First, we integrate three non-linear rectification layers rather than one, which leads to a more discriminative decision function. Second, we reduce the number of parameters by supposing that the three layer 3 × 3 convolution stack has K channels, and the stack is parameterized by $3(3^2K^2) = 27K^2$ weights; in contrast to a single 7 × 7 convolutional layer which requires $7^2K^2 = 49K^2$, which needs **81%** more of parameters.

*4.8 Comparison with Recent Representative Methods*

The classification performance of the proposed method is compared with the most widely cited approaches and the recent representative paradigms such as: MSRC [34], SVM [37], Cond Gauss [38], MSS [39], A-Convnets [4], CNN [5], BCS [40], CNN–SVM [15] and Deep Leaning [16]. Because the SVM method [37] only considers the three-target classification problem, we have run the published online code from [37] to obtain the results on our classification tasks (ten targets, T-72 Variants, and 30° depression angle). All other compared methods do the same classification tasks as ours. Therefore, we directly use the classification accuracies presented in the corresponding papers. As illustrated in Table 12, our proposed method achieves the highest accuracy rates. Fig. 14 shows MSTAR/IU Mixed Targets recognition results based on MF datasets.

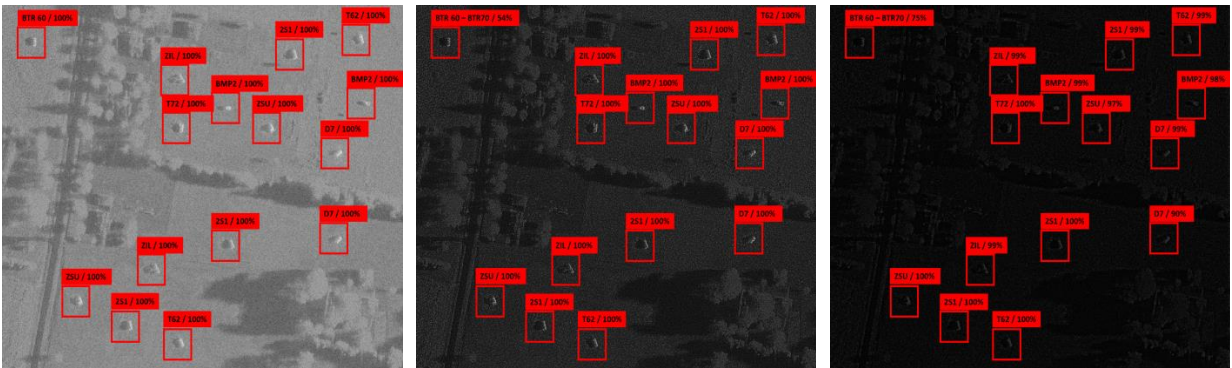

**Fig. 14.** MSTAR/IU Mixed Targets recognition results using different MF datasets

TABLE 13

THE PERFORMANCE COMPARISON BETWEEN THE PROPOSED METHOD AND THE STATE-OF-THE-ART METHODS ON THE MSTAR PUBLIC DATABASE

| Method | MSTAR public mixed target (%) | MSTAR public T-72 Variants (%) | MSTAR with 30° (%) |
|---|---|---|---|
| SVM [37] | 90 | 78.5 | 81 |
| Cond Gauss [38] | 97 | 77.32 | 80 |
| MSRC [34] | 93.6 | - | 98.4 |
| MSS [39] | 96.6 | - | 98.2 |
| CNN [5] | 84.7 | 81.08 | 81.56 |
| A-Convnets [4] | 99.13 | 95.43 | 96.12 |
| BCS [40] | 92.6 | 88.76 | 92.6 |
| Deep Leaning [16] | 92.74 | - | - |
| CNN–SVM [15] | 99.5 | 95.75 | 96.6 |
| **Proposed with softmax** | **99.51** | **96.13** | **98.34** |
| **Proposed with SVM** | **99.70** | **96.98** | **98.87** |

## 5   Conclusion

This paper developed SAR Oriented network to automatically learn very deep features from the data sets, selects adaptive feature layers, and fuses the learned features. The proposed network increases the depth by using more convolutional layers to lessen the speckle noise effect. A dropout technique is supplied to deal with the over-fitting problem due to limited training data sets. In addition, the MCCA algorithm is introduced to combine the selective adaptive layer's features and improves the representations with respect to the correlation objective measured on SAR target images, making the proposed method feasible for SAR image processing. Experimental results on the benchmark of MSTAR database demonstrate the effectiveness of the proposed method compared with the state-of-the-art methods.

39. G. Dong and G. Kuang, "Classification on the monogenic scale space: application to target recognition in sar image," IEEE Transactions on Image Processing, vol. 24, no. 8, pp. 2527–2539, 2015.
40. X. Zhang, J. Qin, and G. Li, "Sar target classification using bayesian compressive sensing with scattering centers features," Progress In Electromagnetics Research, vol. 136, pp. 385–407, 2013.